\documentclass{article}
\usepackage[round]{natbib}

\usepackage{siunitx}
\usepackage{authblk}

\title{Human-Centered Artificial Intelligence and Machine Learning}
\author{Mark O.Riedl}
\affil{School of Interactive Computing\\ Georgia Institute of Technology}

\date{}

\begin{document}

\maketitle

\begin{abstract}
    Humans are increasingly coming into contact with artificial intelligence and machine learning systems. 
    Human-centered artificial intelligence is a perspective on AI and ML that algorithms must be designed with awareness that they are part of a larger system consisting of humans.
    We lay forth an argument that human-centered artificial intelligence can be broken down into two aspects: (1)~AI systems that understand humans from a sociocultural perspective, and (2)~AI systems that help humans understand them. 
    We further argue that issues of social responsibility such as fairness, accountability, interpretability, and transparency.
\end{abstract}

\section{Introduction}

{\em Artificial intelligence} (AI) is the study and design of algorithms that perform tasks or behaviors that a person could reasonably deem to require intelligence if a human were to do it.
Broadly construed, an intelligent system can take many forms: a system designed to be indistinguishable from humans; a speech assistant such as Alexa, Siri, Cortana, or Google Assistant; a self-driving car; a recommender in an online commerce site; or a non-player character in a video game.
We refer to intelligent systems as {\em agents} when they are capable of making some decisions on their own based on given goals. 
{\em Machine learning} (ML) is a particular approach to the design  of intelligent system in which the system adapts its behavior based on data.
It is the success of machine learning algorithms in particular that have lead to recent growth in commercialization of artificial intelligence. 

Humans are increasingly coming into contact with artificial intelligence and machine learning systems. 
At times it is evident, as in the case of Siri, Alexa, Cortana, or Google Assistant. 
It is also evident in the case of self-driving cars or non-player characters in computer games. 
At times it is less evident, as in the case of algorithms that work behind the scenes to recommend products, and approve bank loans. 
Given the potential for intelligent system to impact people's lives, it is important to design intelligent systems with this in mind. 

There is a growing awareness that algorithmic advances to artificial intelligence and machine learning alone are insufficient when considering systems designed to interact with and around humans.
{\em Human-centered artificial intelligence} is a perspective on AI and ML that intelligent systems must be designed with awareness that they are part of a larger system consisting of human stake-holders, such as users, operators, clients, and other people in close proximity. 
Some AI researchers and practitioners have started to use the term human-centered AI to refer to intelligent systems that are designed with social responsibility in mind, such as addressing issues of fairness, accountability, interpretability, and transparency. 
Those are important issues. 
Human-centered AI can encompass more than those issues and in this desiderata we look at the broader scope of what it means to have human-centered artificial intelligence, including factors that underlie our need for fairness, interpretability, and transparency.

At the heart of human-centered AI is the recognition that the way intelligent systems solve problems---especially when using machine learning---is fundamentally {\em alien} to humans without training in computer science or AI. 
We are used to interacting with other people, and we have developed powerful abilities to predict what other people will do any why. 
This is sometimes referred to as {\em theory of mind}---we are able to hypothesize about the actions, beliefs, goals, intentions, and desires of others. 
Unfortunately, our theory of mind breaks down when interacting with intelligent systems, which do not solve problems the way we do and can and come up with unusual or unexpected solutions even when working as intended. 
This is further exacerbated if the intelligent system is a ``black box.'' Black box AI and ML refers to the situations wherein the user cannot even know what algorithms it uses, or that the system is so complicated as to defy easy inspection. 
Regardless of whether an intelligent system is a black box or not, we are seeing more interaction between intelligent systems and people who are not experts in artificial intelligence or computing science. 
How can we design intelligent systems that are capable of helping people understand their decisions? Must we design intelligent systems to solve problems more like humans, or can we augment existing algorithms? What do people need to know about an intelligent system to be able to trust its decisions or be comfortable working with an intelligent system? Can the intelligent system itself convey this information to its users in a meaningful and understandable manner?

Human-centered AI is also in recognition of the fact that humans can be equally inscrutable to intelligent systems. When we think of intelligent systems understanding humans, we mostly think of natural language and speech processing—whether an intelligent system can respond appropriately to utterances. Natural language processing, speech processing, and activity recognition are important challenges in building useful, interactive systems. To be truly effective, AI and ML systems need a theory of mind about humans. Just as we use our commonsense reasoning to interpret and predict the actions of others, intelligent systems can benefit from having commonsense understanding of what people do and why they do it in particular ways; we are unconsciously influenced by our sociocultural beliefs. 
Intelligent systems, however, do not ``grow up'' immersed in a society and culture in the way humans do. These sociocultural beliefs and norms provide a context surrounding everything that people do.
An intelligent system that can model sociocultural beliefs and norms may be able to disambiguate human behavior and make more educated guesses on how to anticipate and respond to human needs. 
At a minimum, intelligent systems that better understand the sociocultural underpinnings of human behavior may be less likely to make mistakes about subjects that people take for granted, making them safer to use and safer for them to be in close proximity to people. 
It may also be possible someday for intelligent systems to evaluate their own behaviors for consistency with ethical norms about fairness.

In the following sections lay out arguments that AI and ML systems that are human-centered (1)~have an understanding of human sociocultural norms as part of a theory of mind about people, and (2)~are capable of producing explanations that non-expert end-users can understand. These are underlying capabilities from which much of our need for AI that contributes to social good stems.

\section{Understanding Humans}

Many artificial intelligence systems that will come into contact with humans will need to understand how humans behave and what they want. This will make them more useful and also safer to use. There are at least two ways in which understanding humans can benefit intelligent systems. First, the intelligent system must infer what a person wants. For the foreseeable future, we will design AI systems that receive their instructions and goals from humans. However, people don’t always say exactly what they mean. Misunderstanding a person's intent can lead to perceived failure. 
Second, going beyond simply failing to understand human speech or written language, consider the fact that perfectly understood instructions can lead to failure if part of the instructions or goals are unstated or implicit.

{\em Commonsense failure goals} occur when an intelligent agent does not achieve the desired result because part of the goal, or the way the goal should have been achieved, is left unstated (this is also referred to as a {\em corrupted goal} or {\em corrupted reward}~\citep{Everitt2017ReinforcementLW}). Why would this happen? 
One reason is that humans are used to communicating with other humans who share common knowledge about how the world works and how to do things. 
It is easy to fail to recognize that computers do not share this common knowledge and can take specifications literally. The failure is not the fault of the AI system---it is the fault of the human operator.

It is trivial to set up commonsense failures in in robotics and autonomous agents. 
Consider the hypothetical example of asking a robot to go to a pharmacy and pick up a prescription drug. Because the human is ill, he or she would like the robot to return as quickly as possible. If the robot goes directly to the pharmacy, goes behind the counter, grabs the drug, and returns home, it will have succeeded and minimized execution time and resources (money). 
We would also say it robbed the pharmacy because it did not participate in the social construct of exchanging money for the product.

One solution to avoiding commonsense goal failures is for intelligent systems to possess commonsense knowledge. 
This can be any knowledge commonly shared by individuals from the same society and culture. 
Commonsense knowledge can be {\em declarative} (e.g., cars drive on the right side of the road) or {\em procedural} (e.g., a waitperson in a restaurant will not bring the bill until it is requested). 
While there have been several efforts to create knowledge bases of declarative commonsense knowledge (CYC~\citep{lenat95}, ConceptNet~\citep{liu04}), these efforts are incomplete and there is a dearth of knowledge readily available on procedural behavioral norms.

There are a number of sources from which intelligent systems might acquire common knowledge, including machine vision applied to cartoons~\citep{Vedantam2015LearningCS}, images~\citep{Sadeghi2015VisKEVK}, and video. 
Unsurprisingly, a lot of commonsense knowledge can be inferred from what people write, including stories, news, and encyclopedias such as Wikipedia~\citep{Trinh2018ASM}. 
Stories and writing can be particularly powerful sources of common knowledge; people write what they know and social and cultural biases and assumptions can come out, from descriptions of the proper procedure for going to a restaurant or wedding to implicit assertions of right and wrong. 
Procedural knowledge in particular can be used by intelligent systems to better provide services to people by predicting their behavior or detect and respond to anomalous behavior.
In the same way that predictive text completion is helpful, predicting broader patterns of daily life can also be helpful.

Combining commonsense procedural knowledge with behavior can yield intelligent agents that are safer. To the extent that it is impossible to enumerate the ``rules'' of society---which is more than just the laws a society has---commonsense procedural knowledge can help intelligent systems and robots follow social conventions. 
Social conventions often exist to help avoid humans avoid conflict with each other, even though they may inconvenience us. 
\citet{harrison:aaai-symbiotic2016} used the pharmacy scenario summarized above to show that intelligent autonomous agents that use written stories as training demonstrations can implicitly learn to avoid behaviors that are socially undesirable. This happens as a side effect because stories are more often than not positive examples of behavior. 
Their system directly addresses the challenge of commonsense goal failures by deriving reward information from following commonly agreed-upon story progressions as close as possible. 
Thus the agent learns not to steal the prescription drugs because most stories refer to exchange of money leading the agent to prefer that behavior even though it is faster and less costly to do the opposite. 
The agent thus never has to be told what ``stealing'' is and why it should be avoided. 

Commonsense knowledge, the procedural form of which can act as a basis for theory of mind for when interacting with humans, can make human-AI interaction more natural.
Even though ML and AI decision-making algorithms operate differently from human decision-making, the behavior of the system is consequently more recognizable to people. 
It also makes interaction with people safer: 
it can reduce commonsense goal failures because the agent fills in an under-specified goal with commonsense procedural details; 
and an agent that acts according to a person’s expectations will innately avoid conflict with a person who is applying their theory of mind of human behavior to intelligent agents. 

\section{AI Systems Helping Humans Understand Them}

Invariably, an intelligent system or autonomous robotic agents will make a mistake, fail, violate an expectation, or perform an action that confuses us. Our natural inclination is to want to ask: ``Why did you do that?'' 
Although people will be responsible for providing goals to intelligent, autonomous systems, the system is responsible for choosing and executing the details.

Neural networks, in particular, are often regarded as {\em un-interpretable}, meaning that it takes a large amount of effort to determine why the systems response to a stimulus is what it is. We often speak of ``opening the black box'' to figure out what was going on inside the autonomous system's decision-making process. A majority of the work to date is on visualizing the representations learned by neural networks (e.g., generating images that activate different parts of a neural network~\citep{Zhang2018VisualIF}) or by tracing the effects of different portions of the input data on output performance. (e.g., removing or masking parts of training data to see how performance is affected~\citep{Ribeiro2016ModelAgnosticIO}). Even AI experts can have a hard time interpreting machine-learned models and this type of work is geared largely toward AI power-users, often for the purposes of debugging and improving a machine learning system.

However, if we want to achieve a vision of autonomous agents and robots being used by end users and operating around people, we must consider non-expert human operators. Non-experts have very different needs when it comes to interacting with autonomous agents and robots. 
Non-expert operators are likely not going to seek a detailed inspection of the inner workings of the system, but is more likely seeking {\em remedy}. 
Remedy is the concept that a user should be able to correct or seek compensation for a perceived failure. 
An intelligent agent did what it thought at the time was the right thing to do only to have been mistaken or to appear to have made a mistake because the behavior violated user expectations. 
The first step in remedy is getting enough information to choose the appropriate remedial course of action. 
Was the failure (or appearance of failure) due to sensor error, effector error, incorrect model, incomplete model, dataset bias, or other cause? 
In many cases the information can be the remedy itself, in the form of an explanation that helps one revise their theory of mind about the agent or as an admission of failure. 
{\em Explanations} are post-hoc descriptions of how a system came to a given conclusion or behavior. 
Explanations can be visual, as in highlighting the portions of sensory input that contributed most to the output~\citep{Selvaraju2017GradCAMVE}, or through natural language processing~\citep{Andreas2017TranslatingN}. 

The question of what makes a good explanation of the behavior of a machine learning system is an open question that has not been explored at depth from a human factors perspective.
One option for natural language explanation is to generate a description of how the algorithm processes sensory input. 
This can be unsatisfactory because algorithms such as neural networks and reinforcement learning defy easy explanation (e.g., ``the action was taken because numerous trials indicate that in situations similar to this the action has the highest likelihood of maximizing future reward''). 

Another option is to take inspiration from how humans respond to the question ``why did you do that?''. 
Humans produce {\em rationales}---explanations after the fact that plausibly justifies their actions. 
People do not know how the exact cascades of neural activation resulted in a decision; we invent a story, consistent with what we know about ourselves and with intent on being as informative as possible. 
In turn, others accept these rationales knowing that they are not absolutely accurate reflections on the cognitive and neural processes that produced the behavior at the time. 
Rationale generation is thus the task of creating an explanation comparable to what a human would say if he or she were performing the behavior that the agent was performing in the same situation. 
\citet{ehsan:iui2019} show that human-like rationales, despite being true reflections of the internal processes of a black-box intelligent system, promote feelings of trust, rapport, familiarity, and comfort in non-experts operating autonomous systems and robots. 

Rationales are generated by first collecting explanations of humans performing a similar task to that of the autonomous system. 
A neural network is trained to translate the internal state of an autonomous agent into the natural language explanations in the corpus. 
This results in automatically generated rationales that read like human rationales, including culturally specific idioms, if present in the corpus. 
This leads to interesting open research questions. 
When is explanation generation by layering one black box on top of another appropriate?
How do incorrect rationales and explanations affect operator perceptions of trust? 
Rationales are probably only one part of the solution---they address a high-level need for explanation and cannot answer specific questions in which an operator asks for elaboration. 
Rationales may be used as one of a set of techniques that meet different levels of need from different types of users.

\section{Conclusions}

With these desiderata, we break human-centered artificial intelligence into two critical capacities: (1) understanding humans, and (2) being able to help humans understand the AI systems. There may be other critical capabilities that this article does not address. However, it seems that many of the attributes we desire in intelligent systems that interact with non-expert users and in systems that are designed for social responsibility can be derived from these two capabilities. For example, there is a growing awareness of the need for fairness and transparency when it comes to deployed AI systems. Fairness is the requirement that all users are treated equally and without prejudice. Right now, we make conscious effort to collect data and build checks into our systems to prevent our systems from prejudicial behavior. An intelligent system that has a model of—and can reason about—social and cultural norms for the population it interacts with can achieve the same effect of fairness and avoid discrimination and prejudice in situations not anticipated by the system's developers. 
Transparency is about providing some means of access to the datasets and workflows inside a deployed AI system to end-users. The ability to help people understand their decisions through explanations or other means accessible to non-experts will provide people with greater sense of trust and make them more willing to continue the use of AI systems. 
Explanations may even be the first step toward remedy, a critical aspect of accountability. 

Human-centered artificial intelligence does not mean that an artificial intelligence or machine learning algorithm must think like a human or be cognitively plausible. 
However, it does recognize the fact that people who are non-experts in artificial intelligence or computing science fall back on a theory of mind designed to facilitate their interaction with other people and draw on sociocultural norms that have emerged to avoid human-human conflict. 
Making intelligent systems human-centered means building the intelligent systems to understand the (often culturally specific) expectations and needs of humans and to help humans understand them in return. 
The pursuit of human-centered artificial intelligence presents a research agenda that will improve our scientific understanding of fundamental artificial intelligence and machine learning while simultaneously supporting the deployment of intelligent products and services that will interact with people in everyday contexts.

\bibliographystyle{plainnat} 
\bibliography{hcai}

\end{document}